\documentclass{sigchi}
\CopyrightYear{2017}
\setcopyright{acmlicensed} 
\doi{http://dx.doi.org/10.475/123_4}
\isbn{123-4567-24-567/08/06}
\conferenceinfo{CHI'18,}{April 21--26, 2018, Montreal, Canada}
\acmPrice{\$15.00}

\usepackage{balance}       
\usepackage{graphics}      
\usepackage[T1]{fontenc}   
\usepackage{txfonts}
\usepackage{mathptmx}
\usepackage[pdflang={en-US},pdftex]{hyperref}
\usepackage{color}
\usepackage{booktabs}
\usepackage{textcomp}
\usepackage{array}
\usepackage{enumitem}

\newcommand{\xhdr}[1]{\vspace{1mm}\noindent{{\it #1}} \vspace{1mm} \ \\}

\usepackage{microtype}        
\usepackage{ccicons}          
\usepackage[colorinlistoftodos]{todonotes}

\def\plaintitle{Analogy Mining for Specific Design Needs}

\def\emptyauthor{}
\def\plainkeywords{Computational analogy, innovation, inspiration, creativity, product dimensions, abstraction, focus, text embedding}

\makeatletter
\def\url@leostyle{%
  \@ifundefined{selectfont}{
    \def\UrlFont{\sf}
  }{
    \def\UrlFont{\small\bf\ttfamily}
  }}
\makeatother
\def\pprw{8.5in}
\def\pprh{11in}

\definecolor{linkColor}{RGB}{6,125,233}
\hypersetup{%
  pdftitle={\plaintitle},
  pdfauthor={\emptyauthor},
  pdfkeywords={\plainkeywords},
  pdfdisplaydoctitle=true, 
  bookmarksnumbered,
  pdfstartview={FitH},
  colorlinks,
  citecolor=black,
  filecolor=black,
  linkcolor=black,
  urlcolor=linkColor,
  breaklinks=true,
  hypertexnames=false
}

\begin{document}

\title{\plaintitle}

\numberofauthors{6}
\author{%
  \alignauthor{Karni Gilon\\
    \affaddr{The Hebrew University of Jerusalem}\\
    \affaddr{Jerusalem, Israel}\\
    \email{karni.gilon@mail.huji.ac.il}}\\
  \alignauthor{Felicia Y Ng\\
    \affaddr{Carnegie Mellon University}\\
    \affaddr{Pittsburgh, PA, United States}\\
    \email{fng@cs.cmu.edu}}\\
  \alignauthor{Joel Chan\\
    \affaddr{Carnegie Mellon University}\\
    \affaddr{Pittsburgh, PA, United States}\\
    \email{joelchuc@cs.cmu.edu}}\\
  \alignauthor{Hila Lifshitz Assaf\\
    \affaddr{New York University}\\
    \affaddr{New York, NY, USA}\\
    \email{h@nyu.edu}}\\
  \alignauthor{Aniket Kittur\\
    \affaddr{Carnegie Mellon University}\\
    \affaddr{Pittsburgh, PA, United States}\\
    \email{nkittur@cs.cmu.edu}}\\
  \alignauthor{Dafna Shahaf\\
    \affaddr{The Hebrew University of Jerusalem}\\
    \affaddr{Jerusalem, Israel}\\
    \email{dshahaf@cs.huji.ac.il}}\\
}

\maketitle

\begin{abstract}
Finding analogical inspirations in distant domains is a powerful way of solving problems. However, as the number of inspirations that could be matched and the dimensions on which that matching could occur grow, it becomes challenging for designers to find inspirations relevant to their needs. Furthermore, designers are often interested in exploring specific aspects of a product-- for example, one designer might be interested in improving the brewing capability of an outdoor coffee maker, while another might wish to optimize for portability.  In this paper we introduce a novel system for targeting analogical search for specific needs. Specifically, we contribute a novel analogical search engine for expressing and abstracting specific design needs that returns more distant yet relevant inspirations than alternate approaches.
%
%
%
%
\end{abstract}
%
%
%
%
%
%
%

\category{H.5.3}{ Group and Organization Interfaces}  

\section{Introduction}
Analogy is a powerful strategy for designing new innovations. For example, Thomas Edison invented the kinetoscope (the precursor to motion picture projectors that are used in theaters today) by working out how to do ``for the eye what the phonograph does for the ear'' \cite{edison_kinetoscope}. The Wright brothers solved a crucial aspect of how to keep their invented aircraft stable during flight by analogy to maintaining balance while riding a bicycle \cite{wing_warping}. More recently, a car mechanic created an innovative new way to assist with difficult childbirth, drawing on an analogy to a party trick for removing a cork stuck in a wine bottle \cite{venema_car_2013}. 

Search engines that could support automatic retrieval of relevant analogies for design problems could significantly increase the rate of innovation and problem solving today. The rise of knowledge databases and repositories on the Internet (e.g., the US Patent Database, Google Scholar, Amazon products, etc.) provides a virtual treasure trove of ideas that could inspire solutions across a variety of domains. Research on creativity and innovation suggests that building on analogous inspirations that are not from the same domain as the source problem is a powerful strategy for generating creative ideas \cite{chan_benefits_2011,gentner1997structure,yu2014distributed}. However, finding useful distant analogies in large databases of textual documents remains challenging for existing machine learning models of document similarity \cite{deerwester_indexing_1990,blei_latent_2003,mikolov_efficient_2013,pennington2014glove}, which are largely dependent on surface features like word overlap.

An additional challenge is that in real world contexts with complex problems, designers are often interested in exploring and abstracting specific aspects of a problem rather than considering the problem as a whole. To illustrate, consider the example of the Wright brothers inventing an airplane. Instead of trying to find an analogy for the entire plane, they regularly found analogies for partial problems they needed to solve, such as steering the wings, or controlling balance during flight \cite {johnsonlaird2002functional}. 
For each identified problem, they then needed to \emph{abstract} key properties of the problem in order to find useful analogs in other domains. In the example of steering the wings, they needed to look beyond some aspects of the wings -- such as the color or particular material -- while keeping in mind other aspects -- such as the semi-rigid frame and need for the wing material to remain taut on the frame. Doing so may have led them to avoid overly general abstractions of ``steering'' that ended up less useful (such as the wings of a bird or the rudder of a ship) and towards more targeted analogical inspirations including the twisting of a cardboard box which drove their final design of warping the wings to steer \cite{wing_warping}.

There are two critical parts of the above example: \textbf{focusing} and \textbf{targeted abstraction}. By \emph{focusing} we mean identifying a particular framing or relation for which we would like to find analogies; here, steering the wings or stabilizing the plane during flight. 
Importantly, analogies specific to one focus may not be relevant to another focus; for example, the problem of keeping the aircraft stable in turbulent air led to a very different analogy of riding a bike, suggesting that small unconscious adjustments of the driver could address shifts in air turbulence \cite{guroff_untold_2016}. 

By \emph{targeted abstraction} we mean choosing the key properties of objects that are important to the core problem (e.g., semi-rigid, thin and flat) while dropping out other less important properties (e.g., color, size). For example, in the steering wings problem, the desired abstraction might be something like ``steer <something that is \textit{semi-rigid}, \textit{thin}, and \textit{flat}>'' 
Targeting the abstraction is necessary in order to avoid finding too many irrelevant matches; for example, previous work has shown that abstracting \textit{all} domain-specific features of the core relational structure of a problem yields less relevant analogies than retaining some \cite{yu_distributed_2016}. 

Many real world solutions similarly require multiple problem framings that would benefit from focusing and targeted abstraction; for example, a coffee maker taken camping may benefit from distant inspirations that make it more lightweight, resistant to weather, a better grinder, or allow the camper to know when to pull it off the fire. Together, focus and targeted abstraction make it possible to find inspirations that are analogically relevant to very specific design needs, without being restricted to inspirations from the same/similar domains.

To address the challenge of targeting analogical search for specific design needs, we present a system 
in which a designer can specify a focus for a given product description, and then abstract that focus beyond its surface features in a \textit{targeted} manner by specifying the key properties of the relations and entities involved that are crucial for understanding the core relational structure. To facilitate expressing this query in a machine-readable way, we leverage a large database of commonsense knowledge (CYC) to provide a set of controlled terms that humans can use to express key properties. Our system then uses this focus-abstracted query to computationally search a corpus of potential inspirations for analogically relevant matches that are tuned to the designer's specific design need. We compare this process to previous state-of-the-art approaches for finding analogies among product descriptions, and find that using our \textbf{Focus-Abstracted} 
queries returns inspirations that are high on both relevance (the results meet the needs of the query) and domain distance (the results are from different domains); 
in contrast, state of the art approaches that operate on the whole document or only on specific keywords from the document, either sacrifice relevance or distance. 
These results have promising implications for creativity-support tools that aim to support designers in solving complex problems through analogy.
\section{Related Work}
\label{section:related}


\subsection{Computational Analogy}

The problem of computational analogy has a long history in artificial intelligence research. Early work focused on devising algorithms for reasoning about analogies between manually created knowledge representations that were rich in relational structure (e.g., predicate calculus representations) \cite{falkenhainer1989structure,gentner1983structure}. While these algorithms achieved impressive human-like accuracy for analogical reasoning, their reliance on well-crafted representations critically limited their applicability to mining analogies amongst databases of free-text documents.

At the same time, much work in machine learning and information retrieval has devised methods for finding documents that are relevant to some query from a user. These methods do not focus on analogy in particular (and certainly not on far analogies): while they differ in the specifics of their methods (e.g., using singular value decomposition, or, more recently, neural networks), in general, they attempt to learn semantic representations of words based on the way that words are statistically \textit{distributed} across word contexts in a large corpus of documents; notable examples include vector-space models like Latent Semantic Indexing \cite{deerwester_indexing_1990}, probabilistic topic modeling approaches like Latent Dirichlet Allocation \cite{blei_latent_2003}, and word embedding models like Word2Vec \cite{mikolov_efficient_2013} and GloVe \cite{pennington2014glove}. The semantic representations produced by these methods are quite useful for finding very specifically relevant documents/results for a query, but are limited in their ability to find matches that are \textit{analogically} related to a query (especially if they do not share domain-specific keywords). 

Recent work by Hope et al \cite{hope2017accelerating} proposes to find analogies among free-text consumer product descriptions by learning to predict an overall representation of a product's purpose (what it is good for) and mechanism (how it works). It uses annotators to mark words related to the purpose/mechanism of the product, and weighs the Glove \cite{pennington2014glove}  values of those words to assign an overall purpose/mechanism representation for each document. It then uses a an artificial neural network model (specifically a bidirectional recurrent neural network, or RNN \cite{bahdanau2014neural}) 
to learn the mapping between the product description's word sequence in GloVe representation and the overall-purpose/mechanism representation captures by the purpose/mechanism annotations. Hope et al showed that they could use these representations to find analogies at a significantly higher rate than comparison state-of-the-art approaches like TF-IDF-weighted GloVe vectors of the documents. 

While promising, as noted above, this approach is designed to find analogies for the \textit{overall} purpose of a given product, and may therefore miss analogies for specific aspects of the product (the specific focus of our paper).

\subsection{Abstraction during Analogy-Finding}
The essence of analogy is matching a seed document with other documents that share its core relational structure \cite{gentner1983structure}; when the analogous documents also have many other details that are very different from the seed document, they are known as far or domain-distant analogies. To find far analogies, it is essential to abstract away these irrelevant details from one's representation of the seed document (and possibly also other documents in the corpus).

Some research has explored how to enable people to construct problem representations that abstract away the surface details of the problem. For example, the WordTree method \cite{linsey_design_2012} has people use the WordNet \cite{miller1995wordnet} lexical ontology to systematically ``walk up'' levels of abstraction for describing the core desired functionality, leading to the possiblity of discovering analogous functions in other domains. For example, a designer who wanted to invent a device to fold laundry for students with very limited fine motor skills might abstract the core function of ``folding'' to ``change surface'', which could lead to analogous inspirations like ``reefing'' (rolling up a portion of a sail in order to reduce its area). Yu et al \cite{yu2014searching} explored how to systematically train crowd workers to convert problem descriptions into an abstracted form that ignored irrelevant surface details. 

Importantly, these abstraction methods do not blindly abstract \textit{all} aspects of a problem description. In many cases, humans exert their judgment to select appropriate levels of abstraction, and also do extensive screening of possible matches based on whether they overlap with key properties/constraints in the original problem context. This is important because analogies that are ``too far'' away can actually lead to \textit{less} creative ideas \cite{chan_best_2015,fu_meaning_2013,goncalves_inspiration_2013}. Yu et al \cite{yu_distributed_2016} recently showed that describing the problem context in abstract terms, but retaining a domain-specific description of its key constraints, enabled crowd workers to find more useful far inspirations than a representation that is abstract on both the problem context and its constraints: for example, the description ``make an object \textit{(abstracted problem context)} that does not tip over easily (\textit{concrete constraint})'' yields more useful inspirations for the problem of making a safe chair for kids, compared to ``make an object \textit{(abstracted problem context)} that is safe (\textit{concrete constraint})'' (which yields inspirations for safety that cannot be applied to chairs).

The insight behind this recent innovation is that abstraction should be \textit{targeted}: rather than completely abstracting away from all the properties of the objects involved in the core relational structure (e.g., the wings in the steering problem for the Wright brothers), it is critical to retain the key properties of the objects that are important for the core relational structure. For example, in order to find inspirations that can suggest useful mechanisms for angling wings to steer a plane (e.g., twisting of a cardboard box), designers need to express to a search engine that they don't care about the color and size of wings, but they \textit{do} care the fact that the wings are \textit{flat, physical objects}, or even that they are composed of materials that respond to \textit{shear forces} in a similar way to cardboard. This insight is consistent with classic cognitive models of analogy (cited above, e.g., \cite{falkenhainer1989structure}), which retain key properties of objects during analogical mapping that are essential to the core relational structure: for example, in the atom/solar-system analogy, the absolute size of the sun/planets vs. nucleus/electron doesn't matter, but the fact that they have mass does.

We build on these insights to explore how we might create a focus-abstracted representation that enables a \textit{computational semantic model} to find more relevant and distant inspirations.

\section{Data Set}
\begin{figure}
\centering
  \includegraphics[width=0.9\columnwidth]{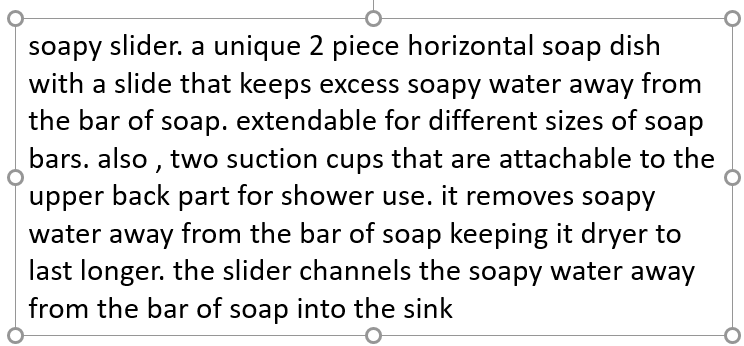}
  \caption{Example: Soapy slider}~\label{fig:soap}
\end{figure}

We use a corpus of product descriptions from \cite{hope2017accelerating}. The products in the corpus are from Quirky.com, an online crowdsourced product innovation website. 
Quirky is useful for our study because it is large (the corpus includes 8500 products) and covers multiple domains, making cross-domain analogies possible. 
Quirky users submit their ideas in free, unstructured (and often colloquial) natural language.
Figure \ref{fig:soap} shows an example submission, demonstrating typical language.

\begin{figure*}[!ht]
\centering
	\includegraphics[width=1.9\columnwidth]{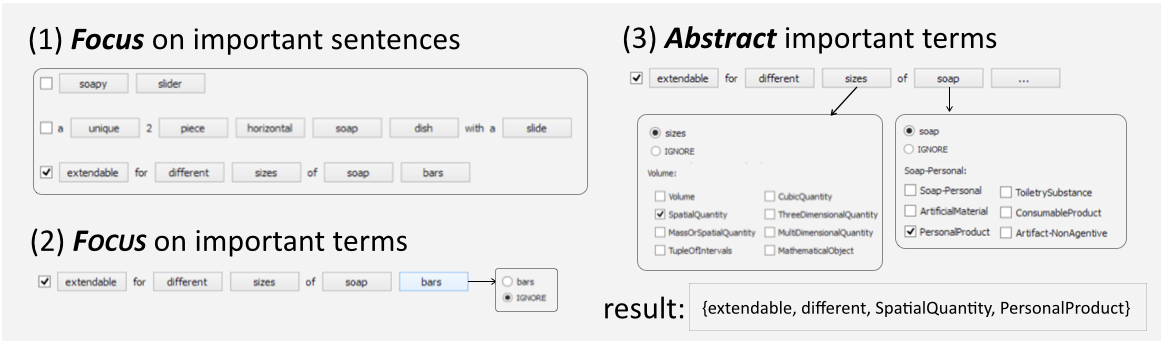}
	\caption{
    Illustration of the process of expressing focus-abstracted queries in our system. Designers express a focus-abstracted query by (1) selecting important sentences in a product description that relate to an intended focus, (2) ignoring irrelevant terms, and (3) replacing important terms (where desired) with appropriate abstracted properties, yielding a sequence of important terms and abstracted properties.} 
    \label{fig:ex-expressing}
\end{figure*}


\section{Method}


When approaching a product different designers may wish to focus on different parts of it. For example, consider the ``Soapy slider'' product in Figure \ref{fig:soap}. One designer (designer A) may wish to explore ways of adjusting to different soap bar sizes while another (designer B) may be interested in removing soapy water from soap bars.
To satisfy their needs, a straightforward approach is for the designers to search the corpus for keyword matches, for example ``change soap size'' for one designer, and ``remove soap water'' for another. Other approaches are mentioned in the Related Work section.

We propose a process that enables the designer to focus on a key need, and then abstract the description to include only the properties of that need that are actually important. For example, Designer A (originally interested in changing the size of soap bars) can indicate that the ``soap'' domain is not important and can be replaced by the more general property of being a ``personal product''. 
The system then finds matches in the corpus using a method based on the state-of-the-art purpose representation engine from \cite{hope2017accelerating}. 
The matches will be analogous products that adjust to different sizes of some personal product, which could hopefully inspire the designer.   

From a system standpoint, the process can be divided into two phases: 1) expressing a focus-abstracted query, and 2) using the query to find analogies in a corpus. 
Below we describe our process in more detail for each phase. 

\subsection{Expressing the focus-abstracted query}
Figure \ref{fig:ex-expressing} shows a worked example of the overall process. We describe each step in turn.

\subsubsection{Step 1: Focus on important sentences}
We assume the designer begins with an existing complete product description. Figure \ref{fig:ex-expressing} shows one such example from the Quirky corpus (the ``Soapy slider'' product). 
The designer selects the sentences most relevant to their need, thus identifying which aspect in the product they wish to further explore. In the ``Soapy slider'' example (Figure \ref{fig:soap}), designer A (focusing on product size adjustments) will choose the sentence ``extendable for different sizes of soap bars.'' (see Step 1 in Figure \ref{fig:ex-expressing}). Designer B (interested in removing liquid from things) will choose the sentence ``it removes soapy water away from the bar of soap keeping it dryer to last longer''.

\subsubsection{Step 2: Focus on important terms}
Important sentences (from Step 1) may still contain terms or domain details that are irrelevant to the intended purpose of the designer. Ignoring irrelevant terms increases the focus of the query on the intended purpose of the designer. It also achieves a form of abstraction (e.g., ignoring irrelevant domain details).

To achieve this function, the interface allows the designer to take any noun, verb, or adjective from the important sentences and mark them with an ``IGNORE'' flag if they are irrelevant to her specific purpose. For example, designer A (who is not interested in \textit{bars} specifically), might choose to IGNORE the term ``bars'' (see Step 2 in Figure \ref{fig:ex-expressing}). Designer B (who is interested specifically in how to remove water from a bar of soap) may IGNORE the term ``last'', which describes the ultimate purpose of keeping the bar of soap dry, but may not be shared by other products that also separate water from objects. 

\begin{figure*}
\centering
	\includegraphics[width=1.75\columnwidth]{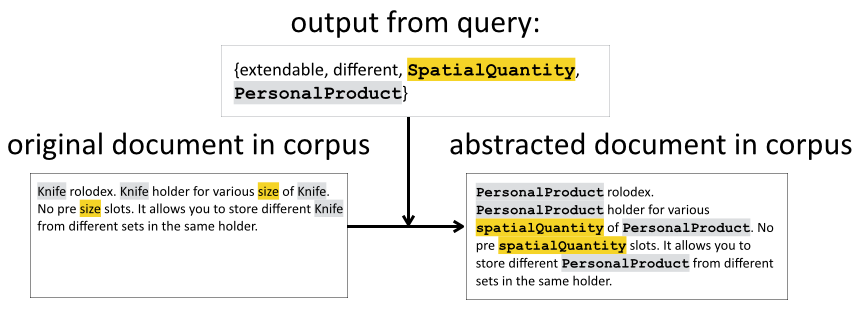}
	\caption{
    Illustration of the process of abstracting the corpus based on a given focus-abstracted query. All terms in each document that match the designer-selected abstracted \textit{properties} (shown in monospace font) are replaced by the matching properties. This brings documents that might be different in domain (e.g., about ``'knives'') but are nevertheless similar at the desired level of abstraction (e.g., PersonalProduct) closer to the focus-abstracted query.} 
    \label{fig:ex-abstracting-corpus}
\end{figure*}

\subsubsection{Step 3: Abstract important terms}
After Step 2, the designer's need are still expressed in the original domain (e.g., soap). In order to find domain distant analogies it is necessary to replace key terms with their appropriate abstractions. 
The designer can abstract a given term by selecting the appropriate abstractions from a list that appears when the term is clicked on. For example, designer A might not be interested in the fact that soap is a ToiletrySubstance, but rather that it's more generally a PersonalProduct, and select that property to abstract the term ``soap'' (see Step 3 in Figure \ref{fig:ex-expressing}).

In designing this component of our system, we faced and dealt with several design challenges:

\begin{itemize}
\item 
\textit{\textbf{Choosing an appropriate knowledge base}}. To find abstractions to show the designers, we explored several knowledge bases. 
Wordnet  \cite{miller1995wordnet} 
is a large English lexical database, including relations like synonym, hypernym and hyponym. Wordnet is lexical and not focused on facts about the world, rendering it less useful for our purpose. In addition, the number of relations it supports is very small. Another alternative we considered is Conceptnet \cite{speer2012conceptnet}. Conceptnet includes knowledge from crowdsourcing and other resources, rendering it very noisy.

We ended up choosing Cyc \cite{Lenat1995cyc} \cite{Lenat1990cyc} as our main Knowledge Base. Cyc is a very large, logic-based knowledge base representing commonsense knowledge. Cyc contains over five hundred thousand terms, seventeen thousand types of relations, and over seven million assertions, i.e., simple facts about the world.  Examples of assertions:  ``\#\$isa \#\$DonaldTrump \#\$UnitedStatesPresident'' (Donald Trump is the US president) and ``\#\$genls \#\$Water \#\$ LiquidTangibleThing'' (liquid is a generalization of water). If a term is missing from Cyc, we resort to WordNet. 

Crucially for our purposes, Cyc contains supersets of terms, which are useful for abstraction. 
For example, ``DomesticatedAnimal'' and ``CanisGenus'' are supersets of ``dog''. ``soap'' may abstract to its supersets ``ToiletrySubstance'',``WaterSolubleStuff'', ``PersonalProduct'', and many others. Another way of looking at it is that soap has the properties of being a water soluble toiletry substance and a personal product. Thus we use the terms Abstractions and Properties interchangeably. 

The level of abstraction controls the distance from the domain, thus allowing the designers to choose far or near analogies. Importantly, the abstractions also give designers control over the \textit{direction} of the abstraction (e.g., ignore all things that are about cleaning, but make sure they share the property of being personal products, or water soluble).
\item 
\textit{\textbf{Dealing with natural language descriptions}}. Quirky product descriptions are written in unstructured natural language. To obtain and display appropriate abstractions for the designers to select from, we first preprocess the corpus and perform part of speech (POS) tagging. We then apply NLTK WordNet morphy \cite{Bird2009NLTK} to get the canonical form of each term (according to its POS), and use this form to query Cyc for its associated properties. For example, we change ``sizes'' to ``size'' before KB lookup. 

\item 
\textit{\textbf{Presenting a manageable set of abstractions to choose from}}. A final design consideration here is the number of abstractions to present to the designer for consideration, since words in Cyc are often associated with many potential abstractions; for example the term ``dog'' has over 100 different abstractions. 
To limit the number, we experimented with filtering by level of abstraction; we found that three abstraction levels appeared to be an optimal cutoff, above which the terms become to general and uninteresting. The abstractions are ordered by their levels, from specific to general. Within each level we considered ordering by measuring property prevalence. If a vast number of  items share a certain property then it is probably too general (e.g., ``Physical Entity''), and will not be useful. If there are too few items, then maybe the property is too specific and less interesting (e.g. ``HairAccessory'' as an abstraction of ``HairBand''). However since the specific abstractions are relatively rare in CYC and since the relevant abstractions normally appear among the first five to ten abstractions we decided not to implement further ordering.  
\end{itemize}

Once the designer selects appropriate abstractions, the expression phase is complete: the end result is a focus-abstracted query derived from the designer's operations on the original product description: unchecked sentences are omitted, words in the IGNORE list are omitted, and the words that were abstracted are replaced by their abstractions. For example, Designer A's selections would yield the following focus-abstracted query: [extendable, different, SpatialQuantity, PersonalProduct] (see Figure \ref{fig:ex-expressing}, bottom). 

\subsection{Finding analogies for the focus-abstracted query}

Now that the designer has expressed their desired focus and abstraction, we use our analogy engine to find analogies from a corpus of potential matches that are tuned for that particular focus (while preserving abstraction). 

The most important step is to re-represent the corpus with the designer's chosen abstractions. Concretely, for each document, we find all terms in it that share the same abstracted properties (in CYC) as those contained in the focus-abstracted query. For example, if the designer abstracted ``soap''  to ``PersonalProduct'' (indicating that it is not the soap they care about, but rather being a personal product),  the engine looks for other terms in the corpus which share the property of being ``PersonalProduct'' (e.g. ``knife'') and abstracts them to ``PersonalProduct'' as well (see Figure \ref{fig:ex-abstracting-corpus}).
The goal of this abstraction step is to ensure that products from different domains that nevertheless share key relations or properties with the focus-abstracted query at the right level of abstraction can be seen as \textit{close} to the query. 


Next, our engine finds matching products in the abstracted corpus. 
We considered several options for the matching. We build on the work of \cite{hope2017accelerating}, which was shown to find good analogies on the same dataset. In short, \cite{hope2017accelerating} takes natural language product descriptions and uses deep learning (specifically, a bidirectional RNN  \cite{bahdanau2014neural}) to learn vector representations for \emph{purpose} (what is this product good for?) and \emph{mechanism} (how does it work?). Given the vector representations, the algorithm of \cite{hope2017accelerating} finds products with similar purpose but different mechanisms, that might serve as analogies. 


We use a similar algorithm for searching for products; however, in our case, since we focus on relevance for a specific abstracted need, we change the algorithm to focus only on finding similar products with respect to purpose. We do this by computing a similarity score 
between the the purpose representation vector for the focus-abstracted query, and the purpose representations for all documents in the abstracted corpus, and selecting the 10 documents with the highest similarity score. 
\section{Evaluation}
Our core hypothesis is that our system is able to find analogies for focused queries, while still retaining the ability to find analogies from distant domains. 

\subsection{Search Scenarios}
We evaluated this hypothesis across a range of focused query scenarios from seeds sampled from the larger corpus of products. The general scenario is that of a designer looking for novel ways to \textit{redesign} some specific aspect of an existing product. This is a common design task that requires creativity, and may especially benefit from distant analogies (in order to maximize novelty), since the existence of many domain features may make it difficult for the designer to think of alternative mechanisms or domains.

We selected 5 seed products from random samples of the corpus, using the following screening criteria:
\begin{itemize}
	\item Is understandable (e.g., grammatical errors and typos do not overly obscure meaning, can actually visualize what it is)
    \item Has at least two well-specified functions/aspects (so we can define distinct focus query scenarios)
    \item Judged to have corresponding inspirations in the corpus
\end{itemize}

Two members of our research team then identified 2 redesign scenarios for each seed. For example, in the ``Soapy slider'' product (which was one of the selected seeds), the two scenarios were ``make the dish compatible with different sizes of soap bars'', and ``keep excess soapy water away from the bar of soap''. We therefore have a total of 10 search scenarios (see Table \ref{tab:scenarios}).

\begin{table}
	\begin{tabular}{|p{3cm}|p{2.25cm}|p{2.25cm}|}
	\hline
		\textbf{\small Seed product} & \textbf{\small Scenario 1} & \textbf{\small Scenario 2} \\ 
        \hline
		\small \textbf{Soapy slider}. Unique 2 piece horizontal soap dish with a slide that keeps excess soapy water away from the bar of soap. & \small Make the dish compatible with different sizes of soap bars & \small Keep excess water away from the bar of soap \\ 
        \hline
		\small \textbf{Camp brew coffee maker}. Light weight all in one coffee grinder and maker for camping and hiking. & \small Tell when something is done cooking & \small Make food and drink outdoors \\ 
        \hline
		\small \textbf{Laundry folding table}. Table that folds down out of the laundry room wall and provides a surface for folding laundry & \small Make compact a pile of flexible, foldable garments & \small Make compact a piece of furniture \\ 
        \hline
		\small \textbf{On/off velcro pocket shoe}. Attached/detached pocket for any shoe. & \small Attach a small pocket to shoe/ankle comfortably \& durably & \small Make the attached pocket inconspicuous \\ 
        \hline
		\small \textbf{The restsack}. Backpack that doubles as an outdoor chair stool. & \small Carry items on the go & \small Provide a portable seat \\
	\hline
    \end{tabular}
    \caption{Seed product descriptions and associated redesign scenarios used for our evaluation experiment. Descriptions shortened.}~\label{tab:scenarios}
\end{table}



\subsection{Constructing the queries}
The research team members who made the scenarios then used our  system to create focus-abstracted queries for each of the scenarios. Figure \ref{fig:ex-expressing} includes one example scenario and focus-abstracted query. Another example (for the ``Soapy slider'' example) is {\textit{RemovingSomething, LiquidTangibleThing, SolidTangibleThing}} for the scenario need ``keep excess soapy water away from the bar of soap''.



\subsection{Measures}
Our goal is to find \textit{relevant} analogies for a specific aspect of a seed product without being constrained to the same domain as the seed product (i.e., retaining the ability to find domain \textit{distant} yet relevant analogies for a focused need). Therefore, we evaluate the \textbf{relevance} and \textbf{domain distance} of each match for its target query. Both measures were obtained by human judgments of the matches. All ratings were performed blind to the method that produced the match: that is, for each scenario, shared matches were combined across the methods, and the method that produced the match was not shown.

\subsubsection{Relevance} 
We operationalized relevance as the degree to which the match meets the needs expressed in the query. Judgment of relevance took into account three factors: 1) the degree to which it shared the key \textit{purpose(s)} expressed in the query (e.g., make compact, adjust), 2) the degree the objects related to the purpose shared the key \textit{properties} of the key objects in the query (e.g., physical size of soap bars), and 3) the degree to which the purpose (and associated mechanism) was explicitly stated (since some products state a function as a desirable property of the product, rather than a function it aims to achieve). This last factor is included because it is easier for a people to notice and use analogies if the mapping to their problem is explicitly described/highlighted \cite{richland_cognitive_2007}. 

The judgment scale was a 5-point Likert-like scale, with the following anchors (developed after multiple rounds of training and piloting with a separate set of data):

\begin{enumerate}
	\setlength\itemsep{0.25em}
	\item[] \textbf{1} = \textit{Matches none of the key functions and object properties in the query}
    \item[] \textbf{2} = \textit{Implicitly matches a few of the key purpose(s), but none of the key object properties in the query}
    \item[] \textbf{3} = \textit{Implicitly or explicitly matches a few of the key purpose(s) AND a few of the key object properties in the query}
    \item[] \textbf{4} = \textit{Implicitly or explicitly matches most of the key purpose(s) AND a most of the key object properties in the query}
    \item[] \textbf{5} = \textit{Explicitly matches most/all of the key purpose(s) and key object properties in the query}
\end{enumerate}

\begin{figure*}[!ht]
\centering
	\includegraphics[width=1.9\columnwidth]{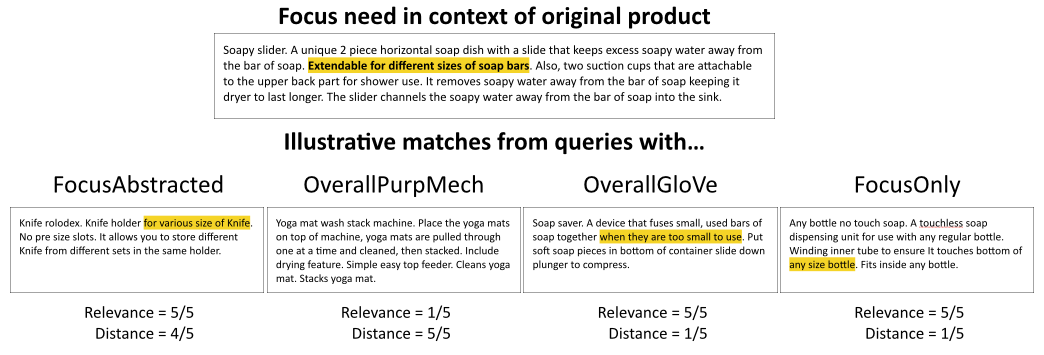}
	\caption{Illustrative matches from each method for the scenario: ``make the dish compatible with different sizes of soap bars''. The abstraction of ``soap'' seems to allow the FocusAbstracted method to ignore the domain difference of knives vs. soap. OverallPurpMech finds a match from a different domain that is analogous in terms of its overall purpose of keeping something clean/dry, but misses the core purpose of adapting to different sizes. In contrast, both OverallGloVe and FocusOnly find a highly relevant match from the same domain.}
    \label{fig:ex-matches}
\end{figure*}

Two members of the research team (who did not directly develop the system) were trained to use the judgments on a separate dataset until they reached good inter-rater agreement, Cronbach's alpha = .82. They then each evaluated half of the matches independently. Examples of low and high relevance-scored matches are shown in Figure \ref{fig:ex-matches}.

\subsubsection{Domain distance}
We operationalized relevance as the degree to which the match shared domain features with the query's seed product. Note that this measure ignores the scenario and instead compares each match with the whole seed product.

The judgment scale was also a 5-point Likert-like scale, ranging from 1 (very similar) to 5 (very different). One issue with this judgment is that many products could be thought of as being in a number of different domains: for example, the 
``camp brew coffee maker'' is about \textit{food/drink/coffee} and \textit{camping/outdoors}. Different raters might weight each feature differently as core/peripheral to the ``domain'' of the product. For this reason, rather than utilizing single ratings from independent raters, each match received a rating from 2 independent judges (members of the research team), and the average of their ratings yielded the distance score for each match. The inter-rater reliability for this measure was good, Cronbach's alpha = .79). Examples of low and high distance-scored matches are shown in Figure \ref{fig:ex-matches}.


\subsection{Experiment Design and Hypotheses}
We compare our method against three other approaches:



\begin{enumerate}
	\item \textbf{OverallPurpMech.} This is the purpose-mechanism method from \cite{hope2017accelerating}. It estimates and matches on \textit{overall} purpose and mechanism vectors. More specifically, we replicate the method from their evaluation experiment, which finds matches for a given seed based on similarity of purpose, and aims to diversify by mechanism. We use this method to find a set of purpose-similar, mechanism-diverse matches for each scenario. The purpose of comparing our system to this method is to determine to what extent we have advanced the state-of-the-art in analogy-finding. 
    \item \textbf{OverallGloVe baseline.} \cite{pennington2014glove} This method approximates the status quo for information-retrieval systems, which tend to operate on the whole document. Each document in the corpus is represented by the average of GloVe vectors for all words (excluding stopwords). We use Glove pre-trained on the Common Crawl dataset (840B tokens, 300d vectors)\footnote{Available here: \url{https://nlp.stanford.edu/projects/glove/}}. We then normalize each document vector, and calculate cosine similarity (which is the same as Euclidean distance in this case) between the resulting vectors for each seed and all other documents, and choose the 10 nearest as matches.
    \item \textbf{FocusOnly baseline.}  This baseline helps tease apart the impact of focusing only versus focusing \textit{and} abstracting (with the knowledge base). For each scenario, we form a focus query as a bag-of-words containing only the words that were abstracted during the process of making the focused-diverse query with our method, i.e., the words themselves instead of their abstractions (stopping at Step 2 in Figure \ref{fig:ex-expressing}). We then calculate the average of the GloVe word vectors for all terms in  the query and compare it with the averaged GloVe vectors of all other products. We again use cosine similarity find the 10 nearest products. 
\end{enumerate}

We therefore have 4 different methods (FocusAbstracted, OverallPurpMech, OverallGloVe, and FocusOnly) of obtaining 10 matches each for 10 different search scenarios. Figure \ref{fig:ex-matches} shows illustrative matches from each othese methods.

We hypothesize that OverallPurpMech will do relatively poorly on relevance (since it is tuned to capture ``the'' overall purpose/mechanism of each document, which might miss the intended focus of a redesign scenario), but well on distance (as it did in \cite{hope2017accelerating}). We hypothesize that OverallGlove will do poorly on relevance \textit{and} distance (since it is neither tuned for focus nor abstraction), and FocusOnly will do well on relevance, but poorly for distance (since it is tuned for focus, but not abstraction). Finally, we hypothesize that our FocusAbstracted method will do well on both relevance \textit{and} distance (since it is tuned for both focus and abstraction).

\begin{figure*}[!ht]
\centering
	\includegraphics[width=1.75\columnwidth]{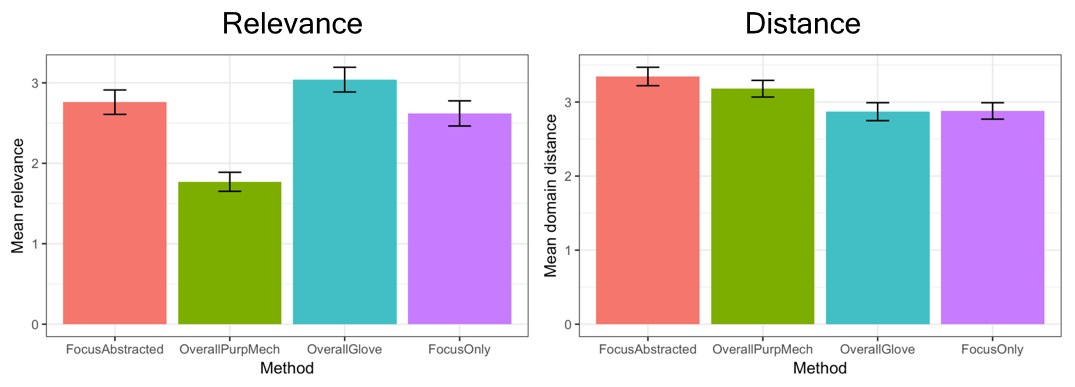}
	\caption{FocusAbstracted matches achieve comparable relevance to FocusOnly and GloVe baselines, and more relevance than OverallPurpMech (left panel) while being more domain distant than FocusOnly and GloVe baseline matches, and equivalently domain distant as OverallPurpMech matches (right panel}
    \label{fig:relevance-distance-by-source}
\end{figure*}

\subsection{Results}

As a first-pass analysis, we note that the methods return almost completely non-overlapping sets of matches for each scenario, giving us 394 unique matches out of 400 total possible unique matches. This initial result suggests that the methods behave quite differently. Indeed, as Figure \ref{fig:ex-matches} illustrates, OverallPurpMech appears to return domain-distant inspirations that are analogous on some purpose of the seed product (though not necessarily the specified purpose), and OverallGlove and FocusOnly appear to return highly relevant inspirations from the same/similar domains, while FocusAbstracted matches appear to be both highly relevant and from distant domains.


\xhdr{FocusAbstracted matches more relevant than OverallPurpMech, and as relevant as OverallGloVe and FocusOnly}
We now turn to formal quantitative tests of our hypotheses. Figure \ref{fig:relevance-distance-by-source} (left panel) shows relevance scores by method, collapsed across scenarios. Using a one-way Analysis of Variance (ANOVA) model with \textbf{method} as the sole between-observations factor, and \textit{matches} as the unit of observation, we find significant differences on the mean relevance score across the methods, F(3,396) = 14.1, $p$ < .01. A follow-up Tukey Honestly Significant Difference (HSD) post-hoc test (to correct for increased chance of false positives due to multiple comparisons) shows that only OverallPurpMech has significantly lower relevance compared to the other methods, $p$ < .01 vs. OverallGloVe, FocusOnly, and FocusAbstracted.

\xhdr{FocusAbstracted matches more domain distant than FocusOnly and OverallGloVe, and as distant as OverallPurpMech}
Figure \ref{fig:relevance-distance-by-source} (right panel) shows distance scores by method, collapsed across scenarios. Again using a one-way ANOVA model with \textbf{method} as the between-observations factor, we find significant differences across the methods on mean distance, F(3,396) = 14.1, $p$ < .01. A follow-up Tukey HSD post-hoc test shows that only FocusAbstracted has significantly higher domain distance compared to OverallGloVe ($p$ < .05) and FocusOnly ($p$ < .05). Despite being numerically more domain distant than OverallGloVe and FocusOnly, the OverallPurpMech method's matches are not significantly more domain distant after the Tukey corrections for multiple comparisons.

\xhdr{Distance of FocusAbstracted matches uncorrelated with relevance, in contrast to OverallPurpMech and FocusOnly}
Finally, we explore the relationship between relevance and domain distance might vary across the methods. Since OverallGloVe tends to match primarily based on surface features, we expect a strong \textit{negative} correlation between relevance and distance, such that relevant matches tend to be \textit{less} domain distant. We expect a similar relationship for FocusOnly (since it operates in a very similar way to OverallGloVe), albeit possibly weaker since it ignores other domain details that were in ignored sentences/terms, but no such relationship for FocusAbstracted and OverallPurpMech (since they are designed to abstract away from domain details. 

Indeed, across all matches from all methods for all scenarios, there is a significant negative correlation between relevance and distance: on average, the more relevant a match is, the closer it is to the domain of the seed product, r = --0.19, 95\% CI = [--0.28, --0.09], $p$ < .01. However, 
the relationship between relevance and distance varies by method. As expected, the relationship is strongest for OverallGloVe matches, r = --0.36 [--0.52, --0.18], followed by FocusOnly, r = --0.22 [--0.40, --0.03], $p$ < .05. In contrast, there is no significant correlation between relevance and distance for either FocusAbstracted, r = --0.09 [--0.28, 0.11], $p$ = 0.38, or OverallPurpMech, r = --0.02 [--0.22, 0.18], $p$ = 0.38. 


\subsection{Case Study}
To give an intuition for what might be driving these quantitative difference, we return to examine 4 illustrative matches for the scenario ``make the dish compatible with different sizes of soap bars'' (shown also in Figure \ref{fig:ex-matches}). OverallPurpMech returns cross-domain matches like a ``yoga mat wash stack machine'', which includes drying and cleaning functions for the yoga mats, which match the overall main purpose of the ``Soapy slider'' product (i.e., keeping the bar of soap dry; in fact, this yoga mat inspiration is relevant for the other ``Soapy slider'' scenario that focuses on this purpose). This illustrates how OverallPurpMech can return interestingly distant but ultimately irrelevant matches if the designer wants to focus on an aspect of a seed product that is different from its main purpose. On the other extreme, OverallGlove and FocusOnly both return many relevant but near matches, like a ``soap saver'' device that fuses small used bars of soap together so they don't slip through the cracks, or a ``touchless soap dispensing unit'' with a winding inner tube that expands to reach inside any size bottle.

In contrast to both of these extremes, our FocusAbstracted method is able to return matches that are both relevant to the focus need \textit{and} domain distant, like a ``knife rolodex'' product that includes multiple slots for different sized knives, or a ``maximizing phone tablet'' (not shown in Figure \ref{fig:ex-matches}), which uses a telescopic frame to adjust to different-sized phones. In both of these cases, our FocusAbstracted method is able to zero in on the idea of adjusting to different ``spatial quantities'', while ignoring differences in the kind of ``personal product'' (e.g., knives, phones) being adjusted to, due to the replacing of the domain-specific terms like knife and phone with abstracted properties that match those of the soap bar.




\section{Discussion}
\subsection{Summary and Implications of Contributions}
In this paper, we sought to design a system that can tune computational analogical search to find relevant \textit{and} distant inspirations for specific design needs. We presented a system that allows designers to focus on a specific aspect of a product description by selecting key terms to form a query, and create a targeted abstraction of those terms by selecting properties from a knowledge base that are important for understanding the core relational structure of the design need. We demonstrated that this focus-abstracted approach led to the retrieval of inspirations that were both relevant and distant, in contrast to alternative state-of-the-art approaches that either sacrificed relevance for distance, or vice versa. Thus, we contribute a promising new method finding distant analogical inspirations for specific design needs.

One specific finding that deserves further discussion is the high performance of the OverallGlove condition in terms of relevance. Our initial prediction was that this condition would perform poorly on relevance, since, like the OverallPurpMech method from \cite{hope2017accelerating}, it operates on the whole product description as opposed to a specific focus query. Cognitive theories of analogy suggest one possible explanation. In particular, some researchers point to the ``kind world hypothesis'' to explain how humans learn abstract concepts: salient surface features (which tend to be shared by things in the same domain) tend to be strongly correlated with structural features. As Gentner \cite{gentner_mechanisms_1989} notes, ``if something looks like a tiger, it is probably a tiger''. One implication of this is that things that are in the same domain likely share many relational features, including purposes. Thus, since OverallGlove is tuned to match based on surface features, it is possible that it found many relevant matches for the specific need simply by finding things in the same domain.

\subsection{Limitations and Future Work}



\subsubsection{Supporting more expressive queries}
We have shown how helpful a focus-abstraction interface can be for a designer wishing to re-design an aspect of a product. However, in our pilot tests of the interface, we noticed that some information needs are still hard to express.
An interesting direction is to explore more expressive queries (by adding more mechanisms to the interface, for example allowing designers to manually add important terms or properties). Improving expressivity might be especially important for domains with highly technical concepts/terms with very specific meanings (e.g., regularization in machine learning) that have poor coverage in existing knowledge bases like Cyc.

\subsubsection{Automatically identifying different purposes in a document}
Our analogy engine calculates a representation of the overall purpose of the focus-abstracted query and all the documents in the corpus. For the abstracted focus description this is a good fit, as the overall purpose is identical to the specific need. For the rest of the corpus the overall purpose comprises several purposes. We expect that automatically dividing  each document to sub-purposes prior to searching for purpose matches would significantly improve them. The usage patterns of our tool can serve as annotated set for learning how to segment the documents. 

\subsubsection{Automatically suggest queries}
Another interesting future direction might be to use information obtained from usage of the tool to learn common focus and abstraction patterns, and suggest focus-abstractions automatically to designers. For example, we might learn that soap dishes, phone cases, and cake-cutters often have a focus-abstracted problem of <expanding to fit objects of different physical sizes>, and suggest these focus-abstractions to other designers creating queries for these (and similar) products.

\subsubsection{Extending to other datasets and domains}
While we have tested our method on Quirky innovations, it would be useful to explore its utility on other corpora. In particular, it would be interesting to test our ideas on a corpus of products from manufacturing companies, which are constantly looking for innovations for improving their products, or even to corpora of research papers. The targeted abstraction approach could be particularly powerful for finding analogies for fields of study where the properties of the objects are critical for determining what makes for useful analogies: for example, as we noticed from our experts in mechanical engineering and materials science, someone working on ways to deform stretchable polymers would not likely benefit from analogies to deformation techniques for concrete, since polymers (by virtue of their material properties) react very differently to physical stresses.

\section{Conclusion}
In this paper, we contribute a novel system for tuning analogical search for specific design needs, consisting of an interface for designers to express their specific needs in abstract terms, and an analogy search engine that uses this focus-abstracted query to find inspirations from a corpus that are both relevant and domain-distant. This work contributes a novel path forward to computational support for mining large databases of potential inspirations on the Web to improve design work.

\bibliographystyle{SIGCHI-Reference-Format}
\bibliography{ref1}

\end{document}